\documentclass[3p,twocolumn,times]{elsarticle}

\usepackage{url}
\usepackage{times}
\usepackage{epsfig}
\usepackage{graphicx}
\usepackage{mathtools}
\usepackage{amsmath}
\usepackage{amssymb}
\usepackage{latexsym}
\usepackage{tabularx}
\usepackage{adjustbox}
\usepackage{booktabs}
\usepackage{multirow}
\usepackage[caption=false]{subfig}
\usepackage{amsbsy}
\usepackage{xfrac}
\usepackage{hhline}
\usepackage{footnote}
\usepackage[table]{xcolor}
\usepackage{amssymb}
\usepackage[hidelinks]{hyperref}

\DeclareMathOperator{\Lagr}{\mathcal{L}}
\DeclareMathOperator*{\argmax}{arg\,max}

\def\etal{\emph{et al.}}

\begin{document}
	\begin{frontmatter}
        \title{Facing the Void: Overcoming Missing Data \\ in Multi-View Imagery}
		
		\author[ufmg]{Gabriel Machado}\ead{gabriel.lucas@dcc.ufmg.br}\corref{cor1}
		\author[stir]{Keiller Nogueira}\ead{keiller.nogueira@stir.ac.uk}
		\author[ufmg]{Matheus B. Pereira}\ead{matheuspereira@dcc.ufmg.br}
		\author[ufmg]{Jefersson A. dos Santos}\ead{jefersson@dcc.ufmg.br}
		
		\address[ufmg]{Departamento de Ci\^{e}ncia da Computa\c{c}\~{a}o, Universidade Federal de Minas Gerais (UFMG), \\Av. Presidente Ant\^{o}nio Carlos, 6627, Belo Horizonte, MG, CEP 31270-901, Brazil}
		\address[stir]{Computing Science and Mathematics, University of Stirling, Stirling, FK9 4LA, Scotland, UK}
		\cortext[cor1]{Corresponding author at +55-31-3409-5860 (phone/fax), Departamento de Ci\^{e}ncia da Computa\c{c}\~{a}o, Universidade Federal de Minas Gerais (UFMG), Av. Presidente Ant\^{o}nio Carlos, 6627, Belo Horizonte, MG, CEP 31270-901, Brazil}
		
		\begin{abstract}

			In some scenarios, a single input image may not be enough to allow the object classification.
			In those cases, it is crucial to explore the complementary information extracted from images presenting the same object from multiple perspectives (or views) in order to enhance the general scene understanding and, consequently, increase the performance.
			However, this task, commonly called multi-view image classification, has a major challenge: missing data.
			In this paper, we propose a novel technique for multi-view image classification robust to this problem.
			The proposed method, based on state-of-the-art deep learning-based approaches and metric learning, can be easily adapted and exploited in other applications and domains.
            A systematic evaluation of the proposed algorithm was conducted using two multi-view aerial-ground datasets with very distinct properties.
            Results show that the proposed algorithm provides improvements in multi-view image classification accuracy when compared to state-of-the-art methods. Code available at \url{https://github.com/Gabriellm2003/remote_sensing_missing_data}.
		\end{abstract}
		
		\begin{keyword}
		    Remote Sensing \sep Image Classification \sep Multi-Modal Machine Learning \sep Metric Learning \sep Cross-View Matching \sep Multi-view Missing Data Completion
		\end{keyword}
	\end{frontmatter}


    
    
    
    
    
    
    
    \section{Introduction} \label{sec:introduction}

Standard image classification tasks are trained by using a single data point as input. 
However, in some cases, using only one input image is not enough to allow its categorization.
One reason for this is the perspective of the object presented in the image, which may not present enough information to allow its identification.
For example, aerial images allow us to observe objects from above, providing information about the general shape and structure of the objects and facilitating the classification of objects such as bridges and streets. On the other hand, ground images give us a closer and frontal view of the object, providing information about fine details and helping the recognition of, for instance, statues and specific facade buildings  (such as schools).

\begin{figure}[ht]
    \centering
    \includegraphics{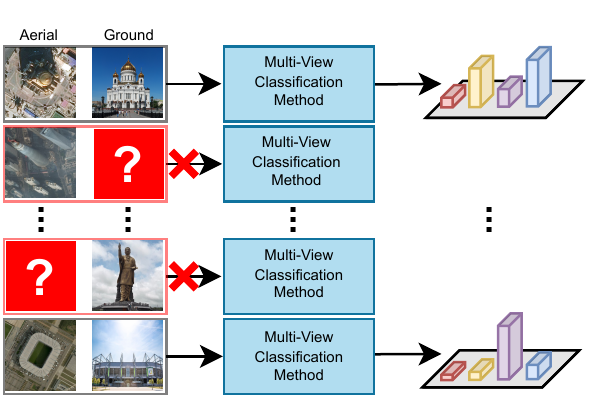}
    \caption {Example of the impact of missing views in the multi-view image classification. A missing view actually represents an instance that can not be used for training, \textit{i.e.}, it represents a severe loss of available information that could be used for the learning process.}
    \label{fig:missing}
\end{figure}

In this context, researchers noticed that it would be essential to exploit the complementary information extracted from images depicting the same object from multiple perspectives (or views) in order to enhance the general scene understanding and, consequently, increase the performance.
This important task, commonly called multi-view image classification, has been successfully explored for distinct applications, including geo-localization~\cite{hoffmann2019model}, mammography analysis~\cite{carneiro2015unregistered}, and land use mapping~\cite{srivastava2019understanding}.
However, although essential and impactful, such task has a major challenge: missing data.
When working with multi-view data, it is really common to have one or more views missing due to malfunction of a sensor, noise, or simply lack of data.
This is even worse when considering that multi-view samples with missing data are often discarded entirely, resulting in a severe loss of available information, as presented in Figure~\ref{fig:missing}.
This issue is especially relevant for domains in which it is difficult to obtain annotated multi-view samples, such as the medical and remote sensing ones~\cite{seeland2021multi}.

In this paper, we propose a novel framework for multi-view image classification capable of dealing with missing data.
Technically, this method can be divided into two parts.
In the first one, a retrieval network, trained using metric learning, receives an input instance and retrieves samples that can be used to fill its missing data gap.
Then, in the second part, information extracted from both the input instance and the top-k retrieved images are further processed using state-of-the-art deep learning-based approaches and then late-fused using standard algorithms in order to perform the final classification. We evaluate the proposed framework on two 2-view (aerial and ground) datasets from the literature, achieving state-of-the-art results. Our methodology, however, can be easily expanded for more complex scenarios with more than two views.



The paper is structured as follows.
Related works are presented in Section~\ref{sec:rw} while the proposed technique is introduced in Section~\ref{sec:methodology}.
Section~\ref{sec:setup} presents the experimental protocol and Section~\ref{sec:results} reports and discusses the obtained results.
Finally, in Section~\ref{sec:conclusion} we conclude the paper and point at promising directions for future work.

    \section{Related Work} \label{sec:rw}

Although several methods have been proposed to tackle multi-view image classification~\cite{seeland2021multi,zhang2021image}, only a few works have investigated and conceived approaches to handle missing data~\cite{zhang2018multi,cai2018deep,choi2019embracenet,srivastava2019understanding,lee2019collagan,aversano2021mic}.

Zhang \etal~\cite{zhang2018multi} proposed a feature-level completion method for missing view of multi-view data.
Technically, their approach first linearly maps multi-view data to a feature-isomorphic subspace, unfolding the shared information from different views.
Then, features of this isomorphic space are used to train a model, which is responsible for retrieving features to represent the missing view and, consequently, completing the multi-view data.
%
In~\cite{choi2019embracenet}, the authors proposed a multi-modal classification framework, called EmbraceNet, that is robust to missing data.
This framework uses a multinomial distribution to select the most relevant features of each view, which 
in order to make this model robust to the partial absence of data, they readjust the multinomial distribution to select features only of the existing modalities.
By doing this, they argue that the missing information due to data loss of a modality can be covered by the other modalities.
%
Finally, Srivastava \etal~\cite{srivastava2019understanding} proposed a two-stream network that extracts discriminative features from aerial and ground images and then combines them for the final classification. 
In order to make their approach robust to missing data, they used these discriminative features to create an embedding space (using Canonical Correlation Analysis (CCA)~\cite{hardoon2004canonical}), which is exploited to retrieve samples that would be similar to the missing data and thus complete the data.

More recently, Generative Adversarial Networks (GANs)~\cite{goodfellow2014generative} have gained popularity in the missing data completion field, due to their ability to generate synthetic samples.
One of the first approaches to investigate the use of GANs for missing data completion was proposed by Cai \etal~\cite{cai2018deep}.
In this work, the authors proposed a conditional GAN~\cite{isola2017image} that takes data from one view as input and generates synthetic samples of the corresponding missing modality.
These data (real and synthetic images) are then used as input to a (discriminator) network which is responsible for the final classification.
Lee \etal~\cite{lee2019collagan} introduced a multi-modal framework, called CollaGAN, for handling missing data imputation.
Precisely, this framework converts the missing data imputation problem into a multi-domain image-to-image translation task.
In this way, a single GAN network can successfully generate the missing data using the remaining (complete) data set.
Another GAN-based work was proposed by Aversano \etal~\cite{aversano2021mic}.
In this work, a multi-branch network jointly encodes data of different modalities/views, generating a common embedding space.
Feature representations of this space are then used in the generation of synthetic data (using the generator network), thus performing data imputation and completing the problematic sample for further processing.

In this work, we propose a multi-view image classification framework that uses a retrieval Convolutional Neural Network (CNN) to handle missing data.
This network was trained using deep metric learning/cross-view matching. Our framework is capable of recovering samples from a different domain just by using the data from the available view. 
Several differences may be pointed out between the proposed approach and the aforementioned works: 
\begin{enumerate}
\item Differently from \cite{isola2017image,lee2019collagan,aversano2021mic}, the proposed technique does not generate synthetic data for missing samples. Compared to retrieval, the task of generating images for one point of view using as input a completely different view of the same object is quite harder~\cite{toker2021coming}, and might introduce more bias. 
\item Our framework handles missing data at the input level, instead of the feature level~\cite{zhang2018multi,choi2019embracenet}
\item Besides, both~\cite{srivastava2019understanding} and our method use an embedding space to retrieve samples, unlike~\cite{srivastava2019understanding}, our method can learn specific features that optimize the class distribution in the embedding space. This is because our model is trained with deep metric learning, instead of using classification features to optimize a CCA model~\cite{hardoon2004canonical}.

\end{enumerate}







    \section{Methodology} \label{sec:methodology}



\begin{figure*}[ht]
    \centering
    \includegraphics[width=\textwidth]{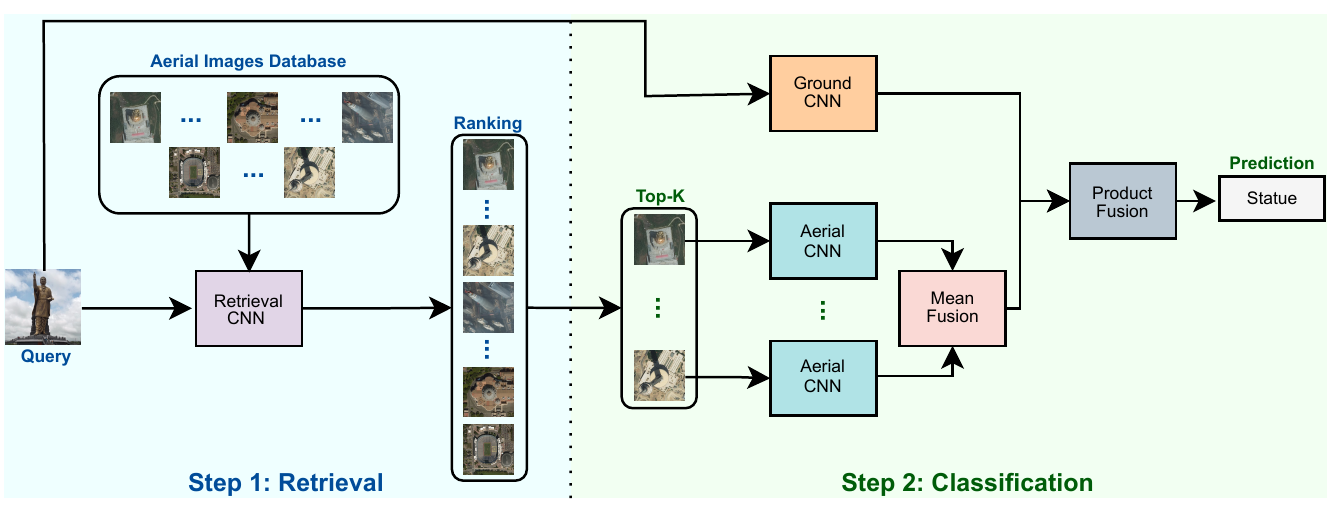}
    \caption {
    Pipeline of the proposed approach applied in a two-view scenario, where aerial images are missing. First, a retrieval network is employed to rank images (of an auxiliary database composed of scenes from the same domain of the missing data) based on their potential to be paired with the input sample with missing data (\textbf{Step 1}).
    Then, the top-$k$ retrieved images and the input data are processed and fused, generating the final classification (\textbf{Step 2}). Following a similar logic, this methodology can be applied to different scenarios. 
    }
    \label{fig:missingdata_inference}
\end{figure*}

The proposed multi-view data classification approach, presented in Figure~\ref{fig:missingdata_inference}, can be splitted into two parts.
In the first one (the retrieval part), a multi-view sample with missing data is processed using a retrieval network responsible for ranking the images (of an auxiliary database composed of scenes of several classes but from the same domain of the missing data) based on their potential to be paired with this input data.
The original image and the generated ranking are then forwarded to the second part, i.e., the multi-view classification.
In this step, retrieved images and original input are processed using their corresponding networks and then fused to produce the final classification. 
Observe that instead of pairing the input image with only the best retrieved image, we select the top-$k$ images in order to alleviate any potential bias and improve generalization.

More details on each of those components, i.e., the retrieval and the classification parts, are presented in the next sections.


\subsection{Retrieval} \label{subsec:m_retrieval}

As introduced, the retrieval network is responsible for recovering images
that could potentially be used as pair for an input example with missing data.

Technically, such a model receives, as input, multi-view image pairs, and is trained using the weighted soft-margin triplet loss~\cite{yu2018correcting}, in which the main objective is to pull these input pairs (commonly called anchor and positive samples) close together whereas pushing dissimilar data (i.e., negative examples) apart.
%
Based on previous works~\cite{crossview2,crossview5,shi2020looking}, the negative instances are mined from the batch using the exhaustive mini-batch strategy~\cite{cv-5}, which proposes to use all other samples (excluding the current anchor and positive pairs) as negative examples. 
Due to this, each batch must have only one image pair per class, in order to ensure that samples from the same class will never be used as negative examples~\cite{carvalho2018cross}.

Formally, suppose the model receives, as input, a batch $B = \{b^1, b^2, ..., b^C \}$ composed of $C$ elements, one of each class, and in which each element is actually a multi-view image pair $b^i = \{ I^i_{v_1}, I^i_{v_2} \}$.
The network first extracts the features for each image pair $F^i = \{ F^i_{v_1}, F^i_{v_2} \}$ and then uses these features to create a matrix of distances $\alpha$ between all images, as presented in Equation~\ref{eq:incidence_matrix}.

\begin{equation}
    \alpha^{i, j} = 2\times(1 - F^{i}_{v_1}F^{j\;\intercal}_{v_2}) \quad \forall~i,j \in C
    \label{eq:incidence_matrix}
\end{equation}

Given the matrix of distances $\alpha$, it is possible to easily calculate the distance between anchors and positive samples ($d_{ap}$ = $\alpha^{i,i}$), and between anchors and negative examples ($d_{an} = \alpha^{i,j} ~~\forall~i \neq j$), and then use those to optimize the model following the aforementioned weighted soft-margin triplet loss~\cite{yu2018correcting}:

\begin{equation}
    \Lagr_{weighted} = \ln{1 + \epsilon^{\gamma(d_{ap} - d_{neg})}}
    \label{eq:soft_margin_triple_loss}
\end{equation}

\noindent where $\gamma$ is a hyper-parameter that controls the loss convergence~\cite{yu2018correcting}.

It is important to highlight two main aspects of the training procedure:
(i) instead of using the exact pair of images as input, random pairs within the same class are employed in order to increase the robustness of the model to missing data (given that, during the testing phase, there will be no corresponding pair to the query image due to missing data);
(ii) for each input image pair, the model is optimized considering one as an anchor and the other as a positive sample and also vice-versa.
By doing this, we allow the model to be prepared for missing data from all possible domains.

During the inference, a multi-view sample with missing data (also called query) is paired with other images (from a previously established database composed of data from the missing domain) that could potentially fill this missing data gap.
All pairs are then processed by the retrieval network and ranked based on their similarity (i.e., on their distance).
All images and the ranking are then forwarded to the classification network to be further processed, as explained in the next Section.

\subsection{Classification} \label{subsec:m_classification}


In the second step of the proposed methodology, the input image with missing data is finally classified.
Towards this, first, the top-$k$ most similar images are selected based on the aforementioned ranking.
Such images are employed to fill the missing data gap of the original query image.
The idea of using the top-$k$ images is to alleviate any potential bias and improve generalization, given that using only one could produce under-representative multi-view pairs whereas using all images could generate instances composed of scenes from different classes.


Query and top-$k$ images are then processed using networks trained specifically for their domains.
Predictions $\sigma$ for the top-$k$ images are fused using the mean operation~\cite{machado2020airound}:

\begin{equation}
    \sigma_{f_2}~=~ \frac{\sum_{i=1}^{k} \sigma_{i}}{k}
    \label{eq:mean}
\end{equation}


Finally, the predictions for the query image ($\sigma_{f_1}$) and the merged predictions of the top-$k$ scenes ($\sigma_{f_2}$) are late-fused using the product operation~\cite{machado2020airound}, thus producing the final classification:

\begin{equation}
    \hat{y}~=~\argmax~ \prod_{i=1}^{2} \sigma_{f_i}
    \label{eq:prod}
\end{equation}

Observe that both mean and product fusion strategies were selected based on previous works~\cite{surveymultimodal,machado2020airound}.
\subsection{Implementation Details} \label{subsec:impl}

The architecture of the retrieval network consists of a two-stream encoder, in which each encoder is actually a pre-trained SwAV~\cite{caron2020unsupervised} model with ResNet-50~\cite{resnet}.
Precisely, this architecture has no classification layers and is trained using the aforesaid weighted soft-margin triplet loss~\cite{yu2018correcting}.

For the inference, instead of using the retrieval network to extract features for all database images for every input query (a costly and time-consuming process), we extract, save, and reuse the features of those database images in order to speed up the testing phase.

For the classification part, we evaluated different Convolutional Networks (VGG~\cite{vgg}, DenseNet~\cite{densenet}, and SKNet~\cite{sknet}).
More details about this, as well as about the employed hyper-parameters, can be seen in Section~\ref{subsec:protocol}.

Lastly, a few adjustments to the methodology can be made to apply our framework to scenarios with N views (with one of them missing). 
To maintain the use of weighted soft-margin triplet loss and consider all the features from the available views, Equation\ref{eq:incidence_matrix} could use the average of these features to compute the distance against samples from the missing view. 
Also, the architecture of the retrieval network must be changed to an N-stream encoder to support N-views inputs. 
Finally, for the classification step, it is necessary to train N classifiers (one per view), and Equation~\ref{eq:prod} must be adjusted to late fuse the prediction scores of all N views (instead of 2).
    \section{Experimental Setup} \label{sec:setup}

In this section, we describe the experimental setup used for the
experiments.
Section~\ref{subsec:datasets} presents the datasets whereas Section~\ref{subsec:protocol} describes the experimental protocol.
Finally, baselines are described in Section~\ref{subsec:baseline}.

\subsection{Datasets} \label{subsec:datasets}

Two multi-view datasets with very distinct properties were used for the experiments in order to better evaluate the effectiveness of the proposed approach.
The first dataset, called AiRound~\cite{machado2020airound}, is composed of 11,753 images divided into 11 classes: airport, bridge, church, forest, lake, park, river, skyscraper, stadium, statue, and tower.
Some examples of these classes are presented in Figure~\ref{fig:airound}.
The second one, named CV-BrCT~\cite{machado2020airound}, comprises approximately 24k pairs of images unevenly split into 7 urban classes: apartment, house, industrial, parking lot, religious, school, store.
Samples of these classes are presented in Figure~\ref{fig:cv_brct}.

For both datasets, each multi-view sample is composed of a ground and an aerial perspective.
Images for the former domain were collected from different sources (such as Google Images, Google Places, Google Street View) and have varying resolutions whereas scenes for the latter perspective were collected using Google Maps and have a fixed resolution of $500 \times 500$ pixels.

\newcommand{\exImages}{0.08}
\begin{figure*}[ht]
	\centering
	\subfloat[Airport]{
	    \begin{tabular}[b]{@{}c@{}}%
    		\includegraphics[width=\exImages\textwidth]{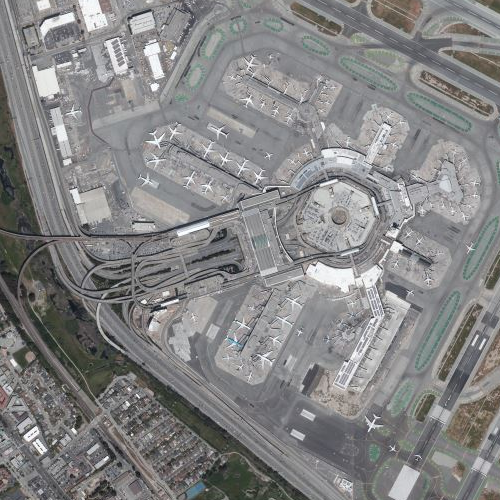} \\
            \includegraphics[width=\exImages\textwidth, height=\exImages\textwidth]{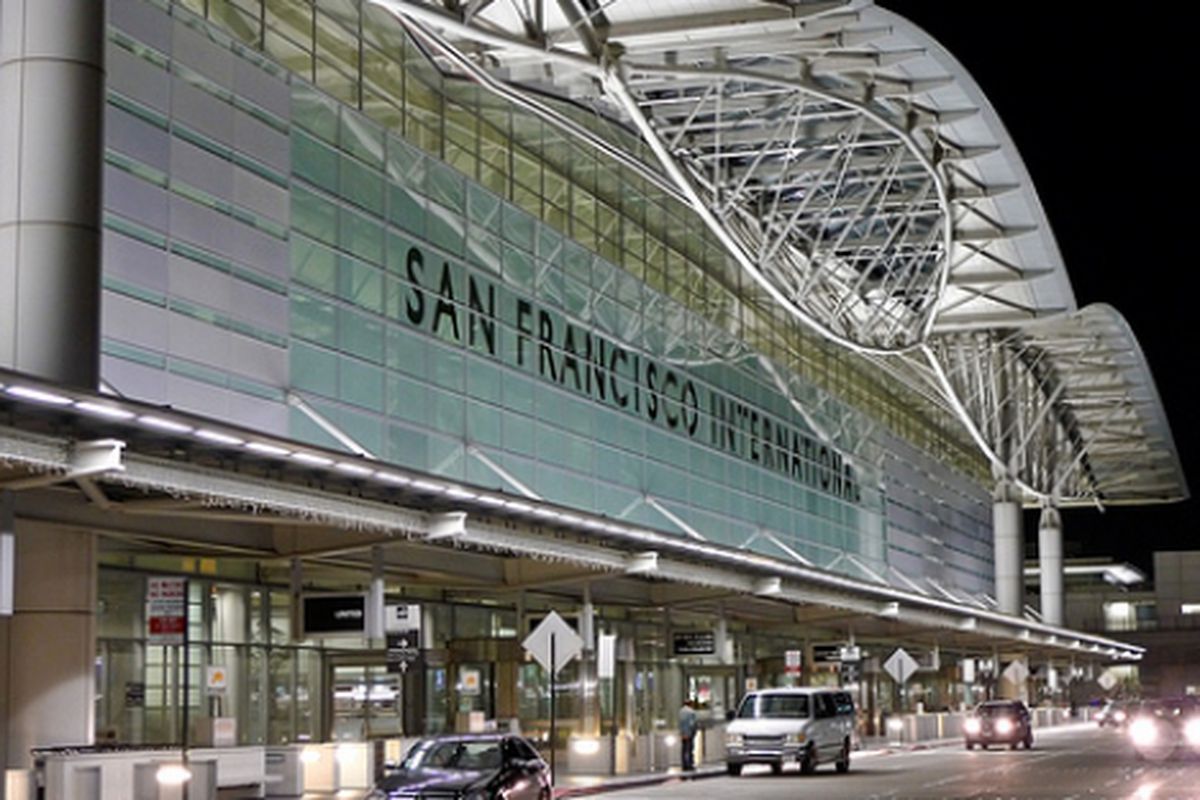}
        \end{tabular}
	}
	\subfloat[Bridge]{
    	\begin{tabular}[b]{@{}c@{}}%
    		\includegraphics[width=\exImages\textwidth]{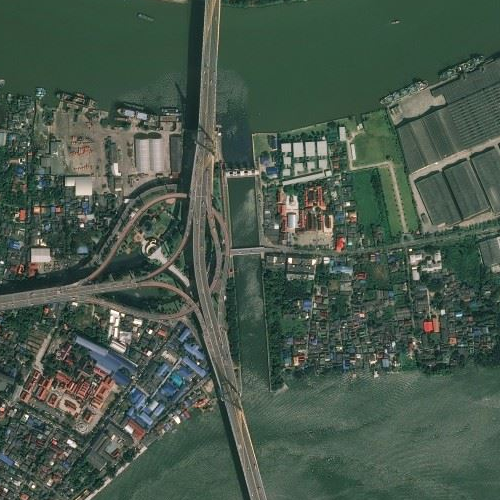} \\
    		\includegraphics[width=\exImages\textwidth, height=\exImages\textwidth]{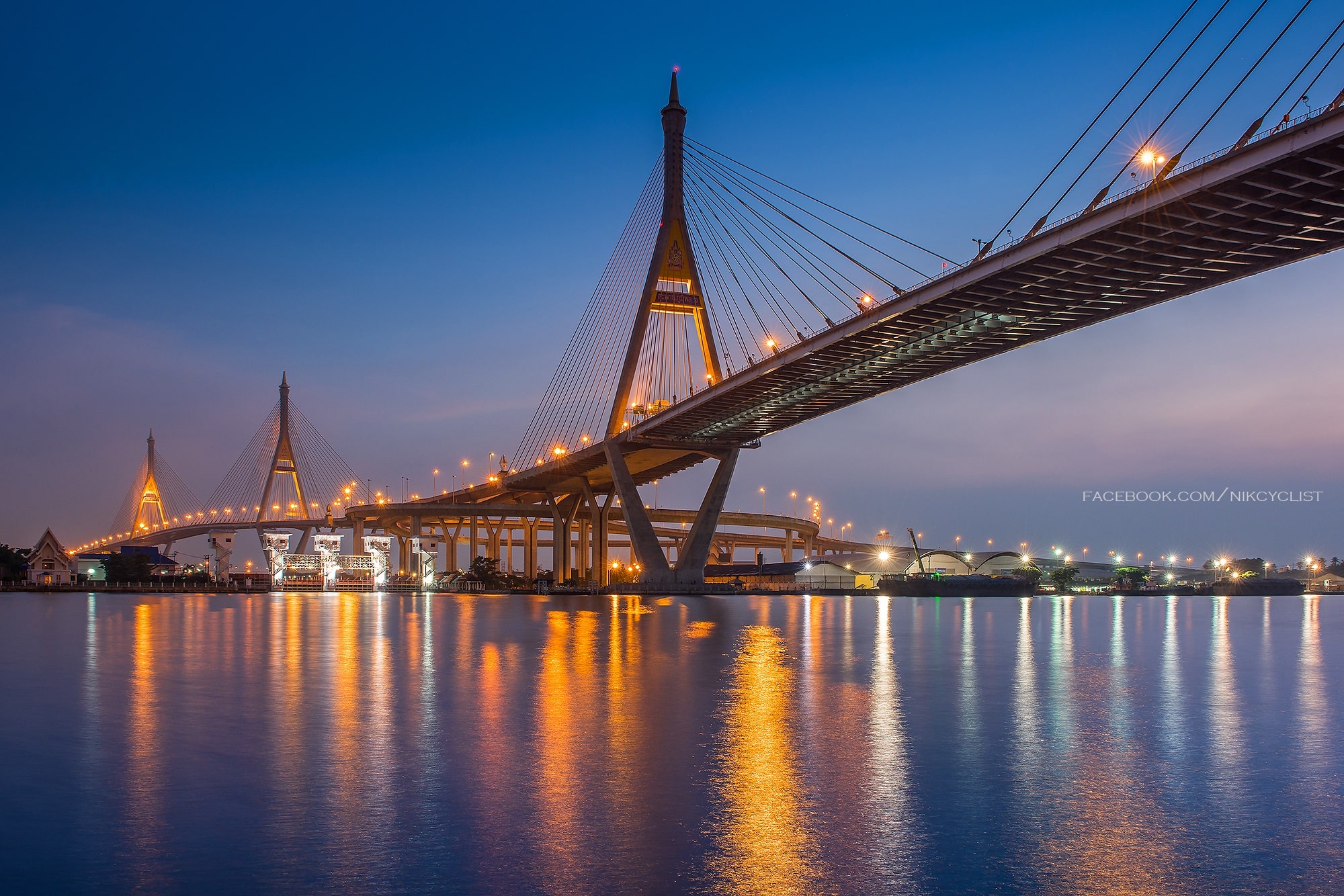}
        \end{tabular}
	}
	\subfloat[Church]{
    	\begin{tabular}[b]{@{}c@{}}%
    		\includegraphics[width=\exImages\textwidth]{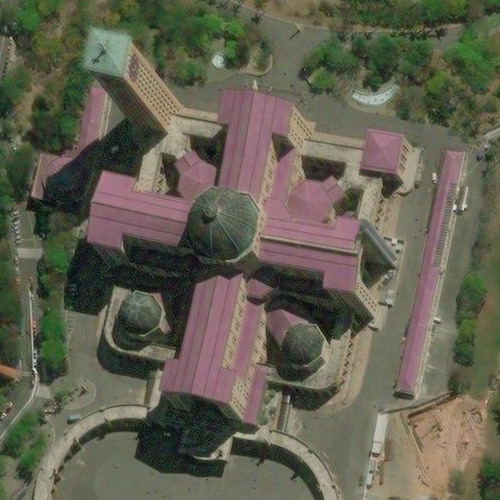} \\
    		\includegraphics[width=\exImages\textwidth, height=\exImages\textwidth]{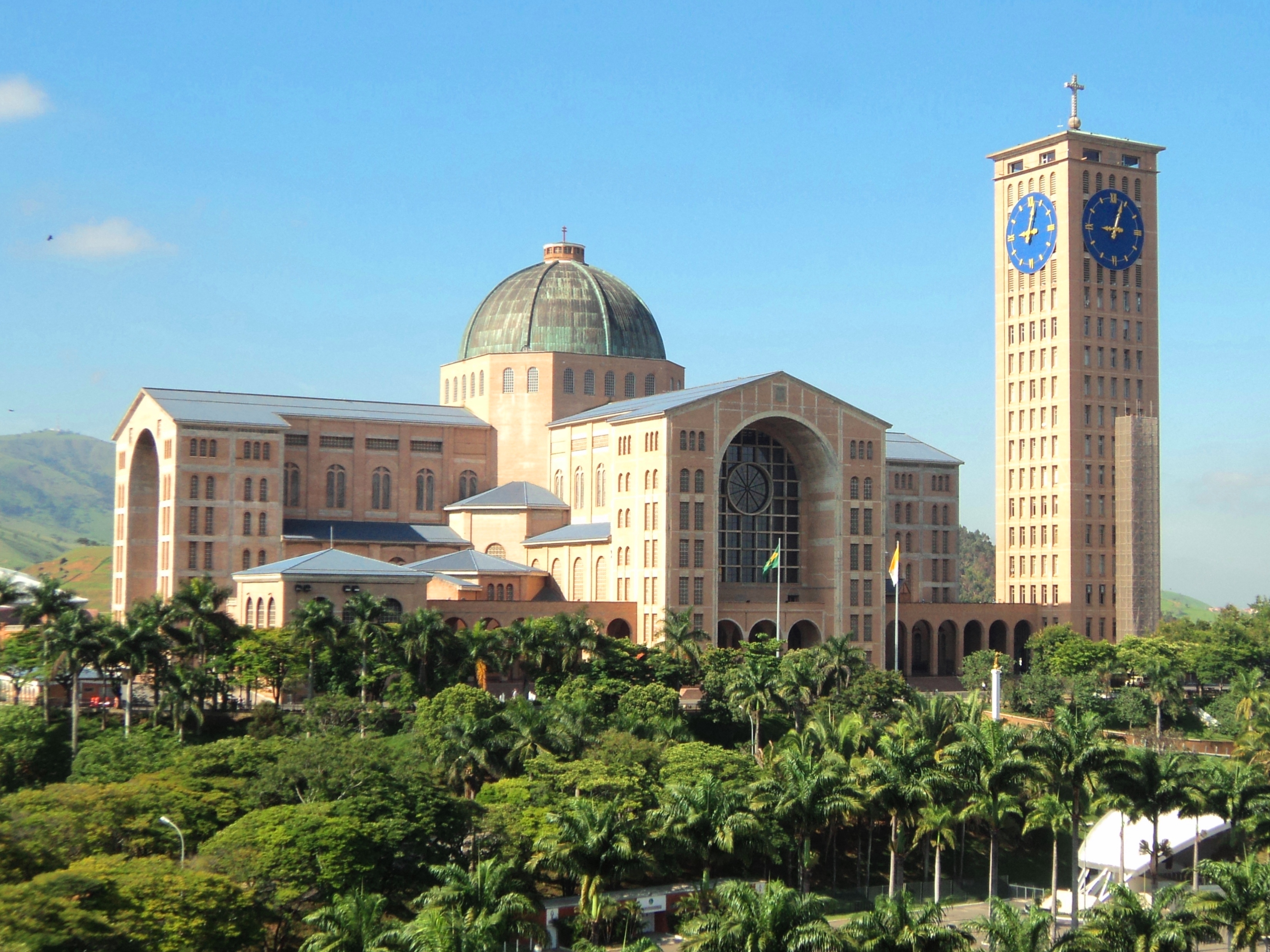}
        \end{tabular}
	}
	\subfloat[Forest]{
    	\begin{tabular}[b]{@{}c@{}}%
    		\includegraphics[width=\exImages\textwidth]{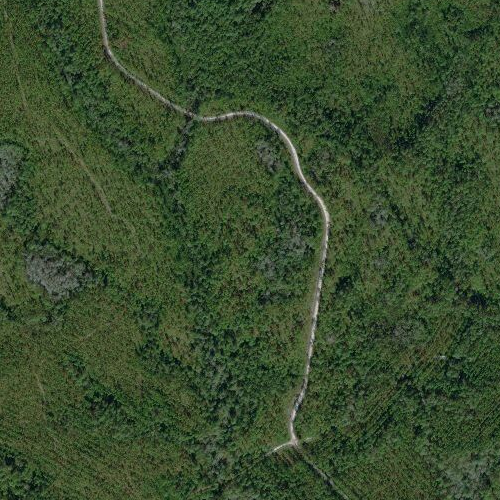} \\
    		\includegraphics[width=\exImages\textwidth, height=\exImages\textwidth]{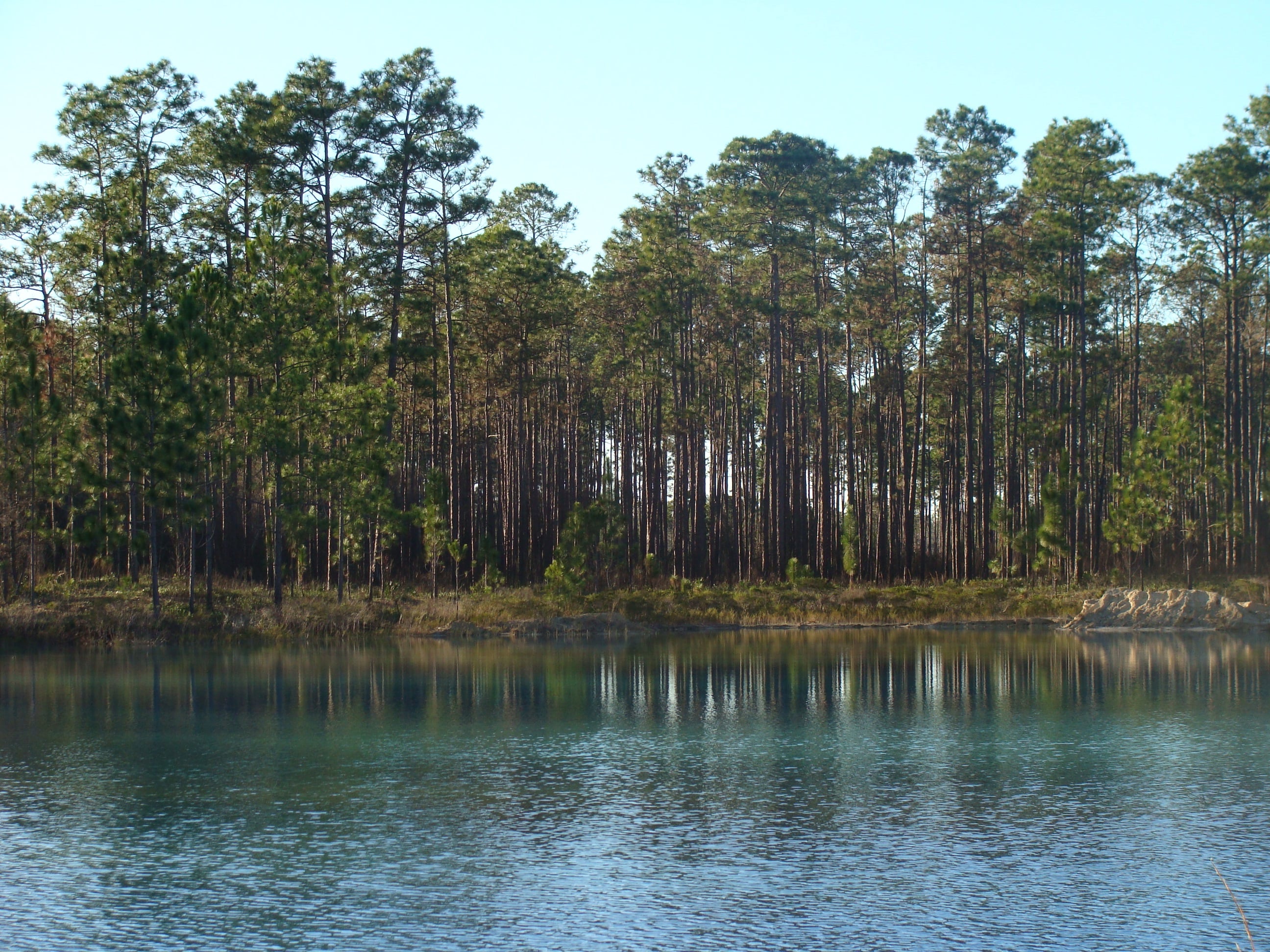}
        \end{tabular}
	}
	\subfloat[Lake]{
    	\begin{tabular}[b]{@{}c@{}}%
    		\includegraphics[width=\exImages\textwidth]{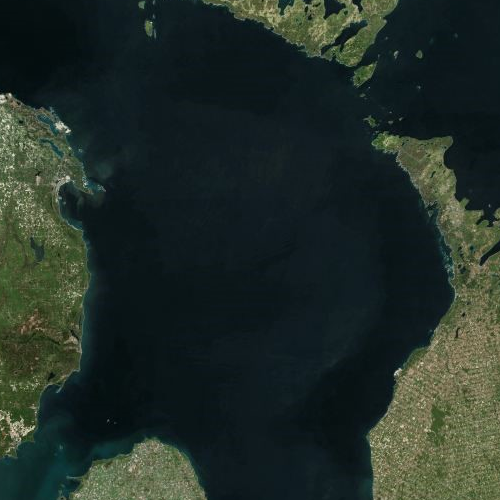} \\
    		\includegraphics[width=\exImages\textwidth, height=\exImages\textwidth]{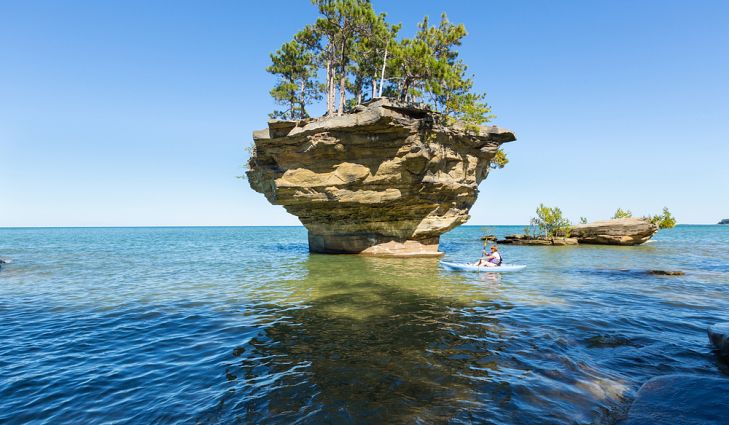}
        \end{tabular}
	}
	\subfloat[Park]{
    	\begin{tabular}[b]{@{}c@{}}%
    		\includegraphics[width=\exImages\textwidth]{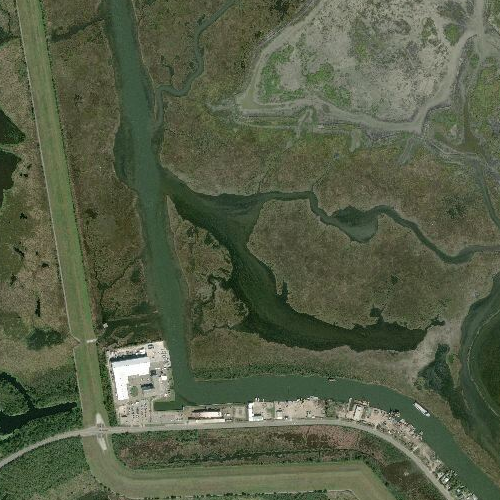} \\
    		\includegraphics[width=\exImages\textwidth, height=\exImages\textwidth]{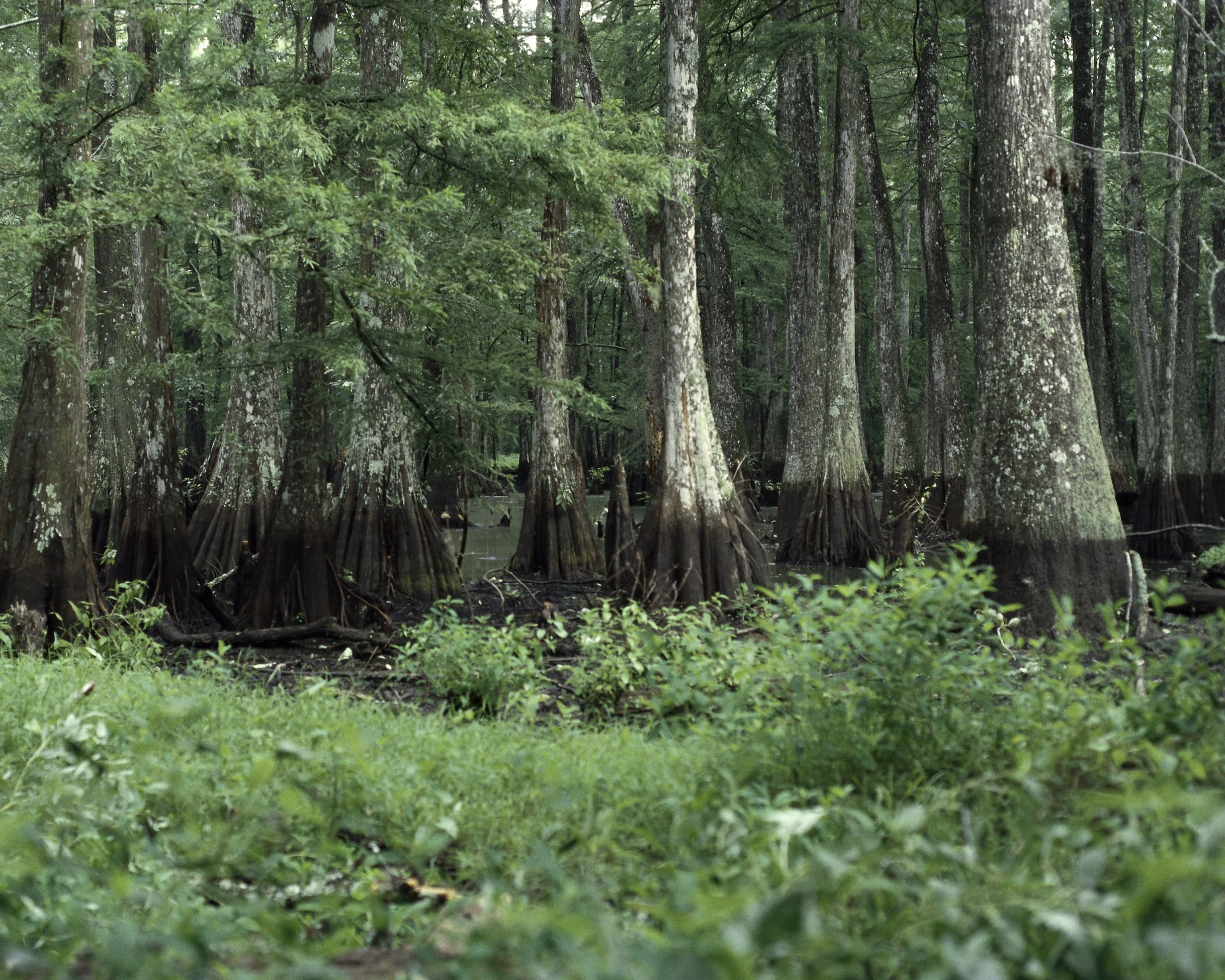}
        \end{tabular}
	}
	\subfloat[River]{
    	\begin{tabular}[b]{@{}c@{}}%
    		\includegraphics[width=\exImages\textwidth]{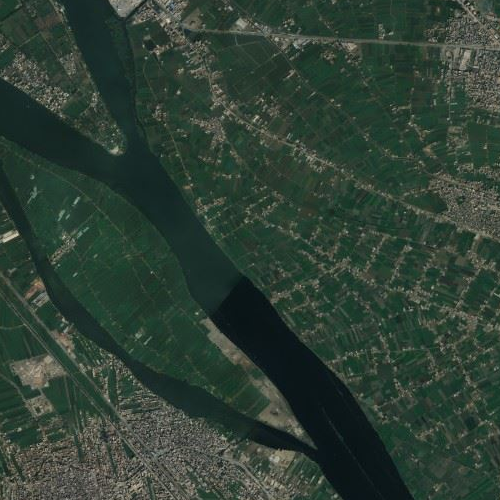} \\
    		\includegraphics[width=\exImages\textwidth, height=\exImages\textwidth]{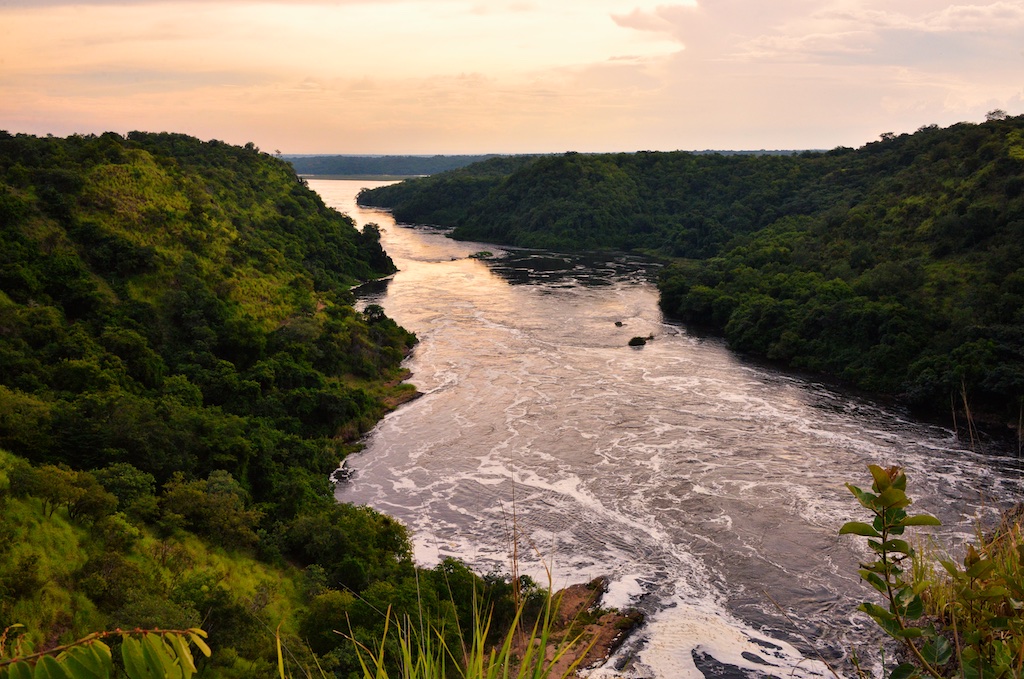}
        \end{tabular}
	}
	\subfloat[Skyscr.]{
    	\begin{tabular}[b]{@{}c@{}}%
    		\includegraphics[width=\exImages\textwidth]{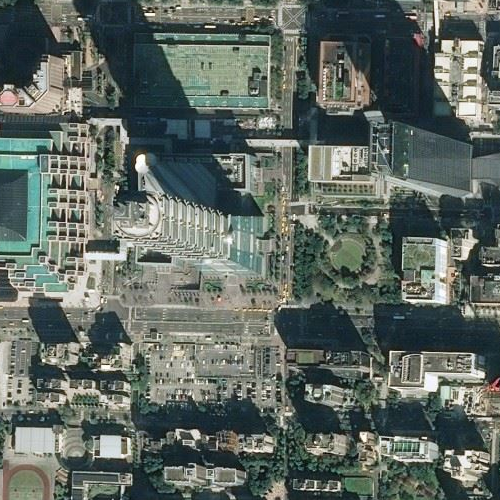} \\
    		\includegraphics[width=\exImages\textwidth, height=\exImages\textwidth]{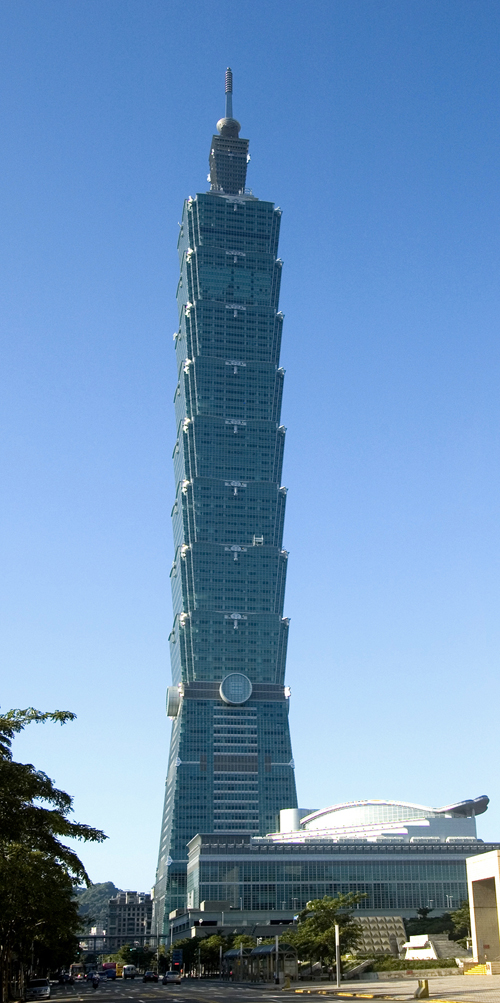}
        \end{tabular}
	}
	\subfloat[Stadium]{
    	\begin{tabular}[b]{@{}c@{}}%
    		\includegraphics[width=\exImages\textwidth]{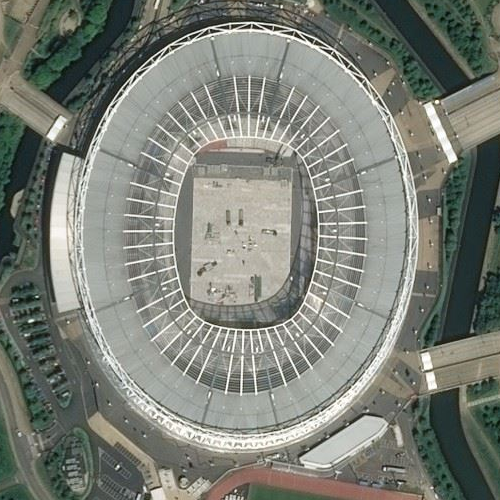} \\
    		\includegraphics[width=\exImages\textwidth, height=\exImages\textwidth]{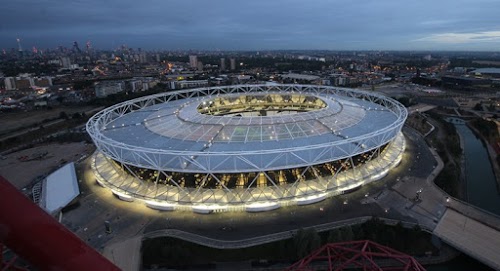}
        \end{tabular}
	}
	\subfloat[Statue]{
    	\begin{tabular}[b]{@{}c@{}}%
    		\includegraphics[width=\exImages\textwidth]{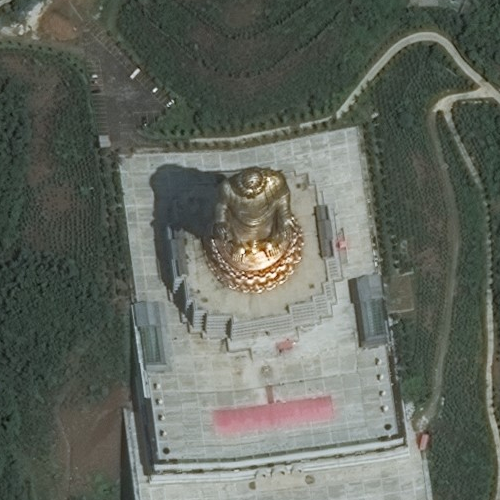} \\
    		\includegraphics[width=\exImages\textwidth, height=\exImages\textwidth]{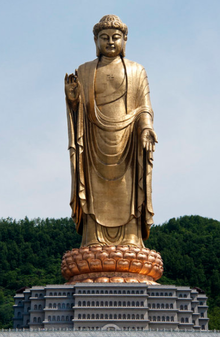}
        \end{tabular}
	}
	\subfloat[Tower]{
    	\begin{tabular}[b]{@{}c@{}}%
    		\includegraphics[width=\exImages\textwidth]{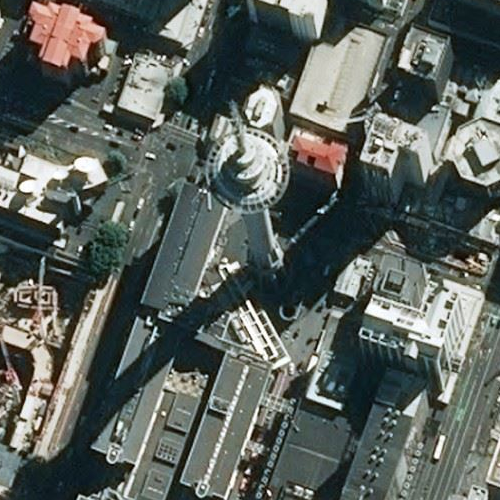} \\
    		\includegraphics[width=\exImages\textwidth, height=\exImages\textwidth]{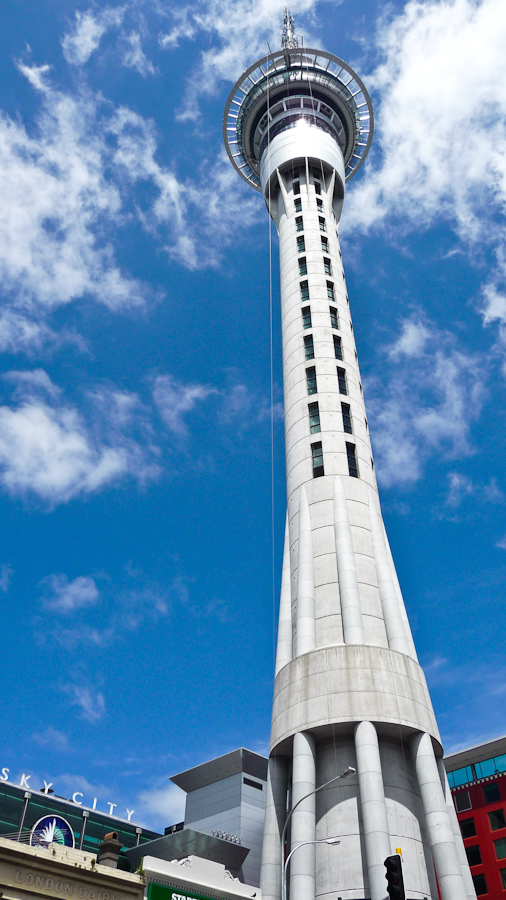}
        \end{tabular}
	}
	\caption{Examples of the AiRound Dataset.}
	\label{fig:airound}
\end{figure*}

\newcommand{\exCV}{0.09}
\begin{figure}[ht]
	\centering
	\subfloat[Apartment]{
	    \begin{tabular}[b]{@{}c@{}}%
    		\includegraphics[width=\exCV\textwidth]{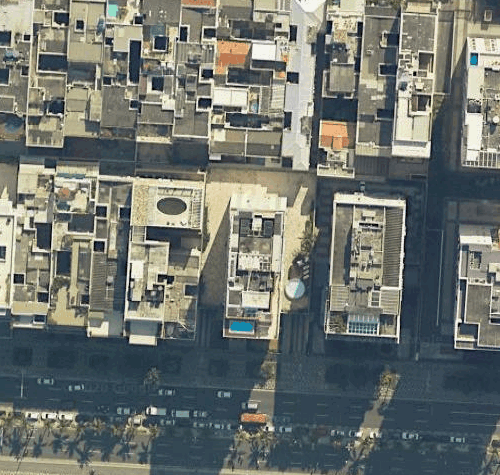} \\
            \includegraphics[width=\exCV\textwidth, height=\exCV\textwidth]{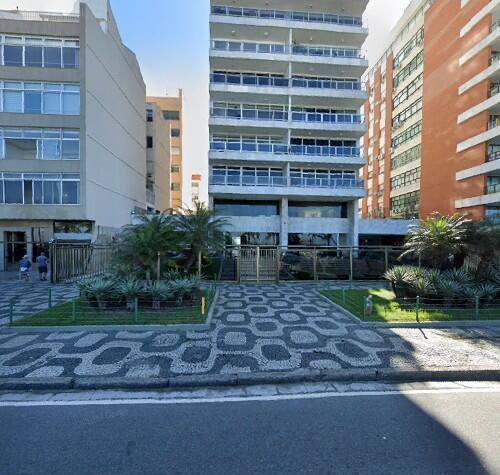}
        \end{tabular}
	}
	\subfloat[House]{
    	\begin{tabular}[b]{@{}c@{}}%
    		\includegraphics[width=\exCV\textwidth]{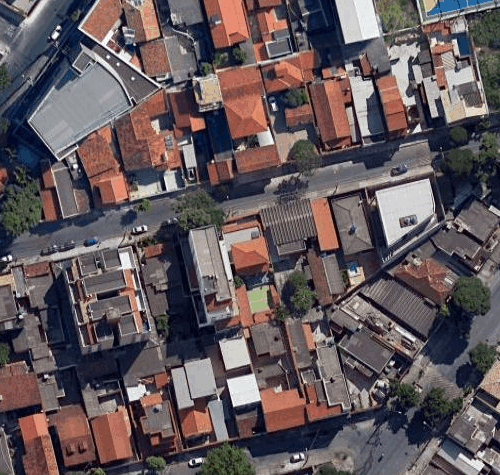} \\
    		\includegraphics[width=\exCV\textwidth, height=\exCV\textwidth]{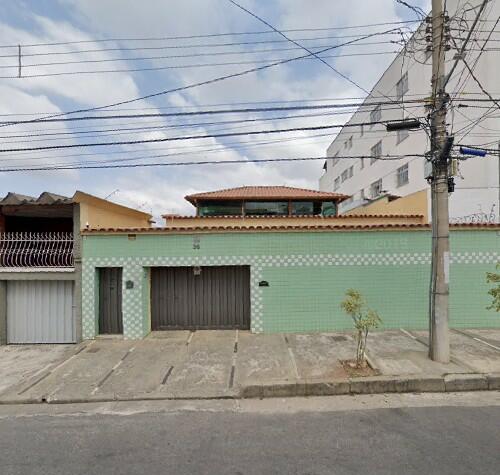}
        \end{tabular}
	}
	\subfloat[Industrial]{
    	\begin{tabular}[b]{@{}c@{}}%
    		\includegraphics[width=\exCV\textwidth]{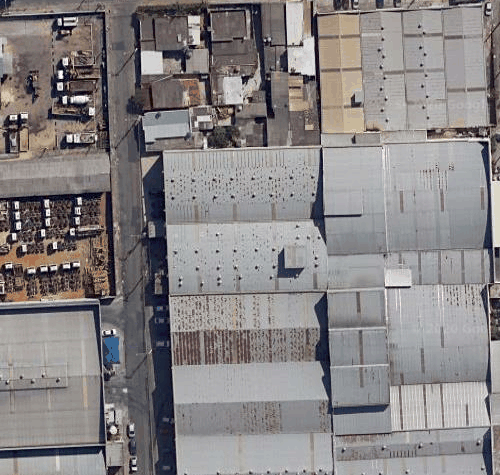} \\
    		\includegraphics[width=\exCV\textwidth, height=\exCV\textwidth]{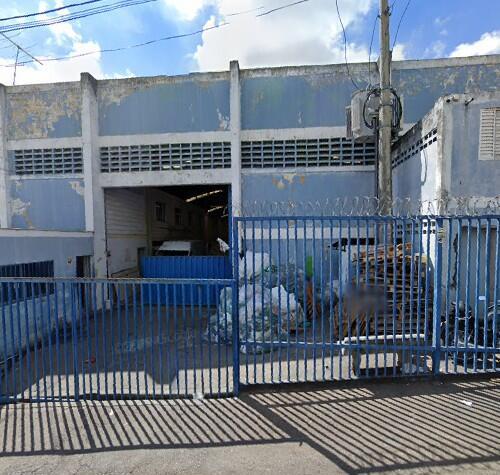}
        \end{tabular}
	}
	\subfloat[Parking]{
    	\begin{tabular}[b]{@{}c@{}}%
    		\includegraphics[width=\exCV\textwidth]{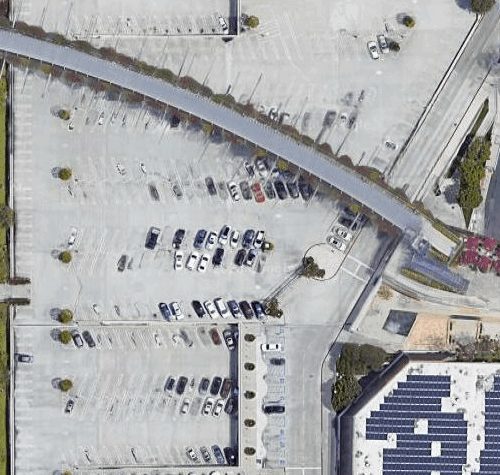} \\
    		\includegraphics[width=\exCV\textwidth, height=\exCV\textwidth]{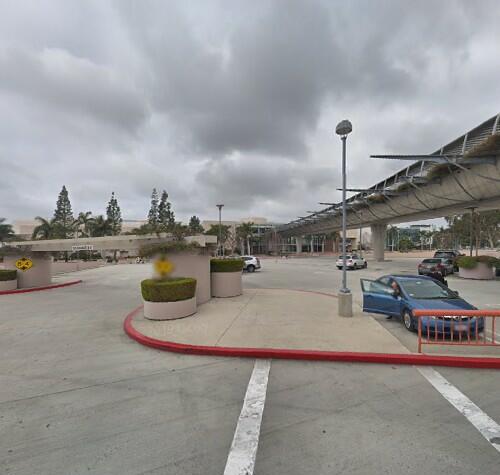}
        \end{tabular}
	} \\
	\subfloat[Religious]{
    	\begin{tabular}[b]{@{}c@{}}%
    		\includegraphics[width=\exCV\textwidth]{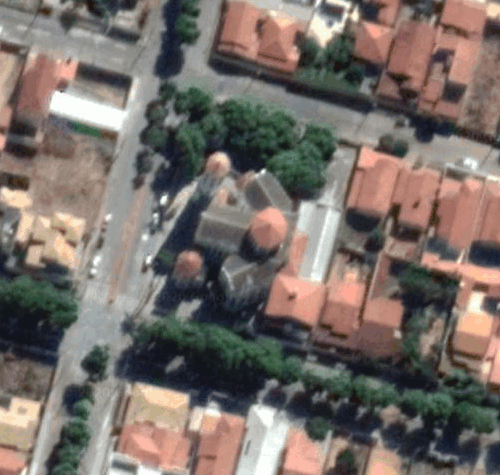} \\
    		\includegraphics[width=\exCV\textwidth, height=\exCV\textwidth]{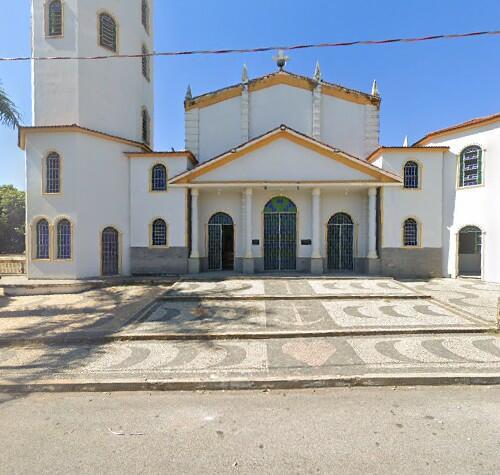}
        \end{tabular}
	}
	\subfloat[School]{
    	\begin{tabular}[b]{@{}c@{}}%
    		\includegraphics[width=\exCV\textwidth]{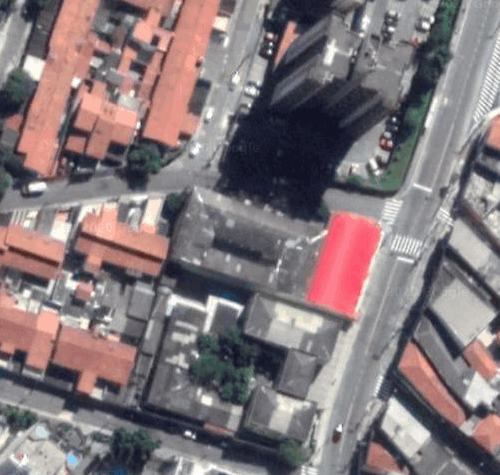} \\
    		\includegraphics[width=\exCV\textwidth, height=\exCV\textwidth]{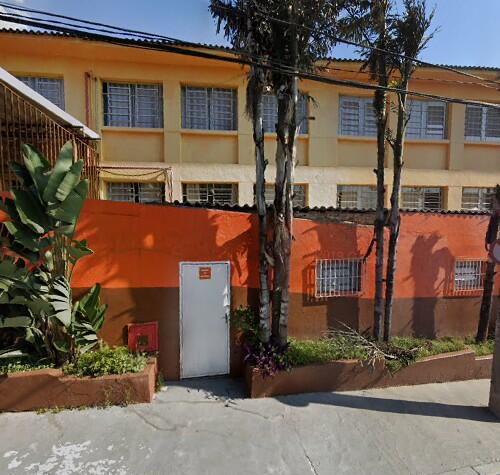}
        \end{tabular}
	}
	\subfloat[Store]{
    	\begin{tabular}[b]{@{}c@{}}%
    		\includegraphics[width=\exCV\textwidth]{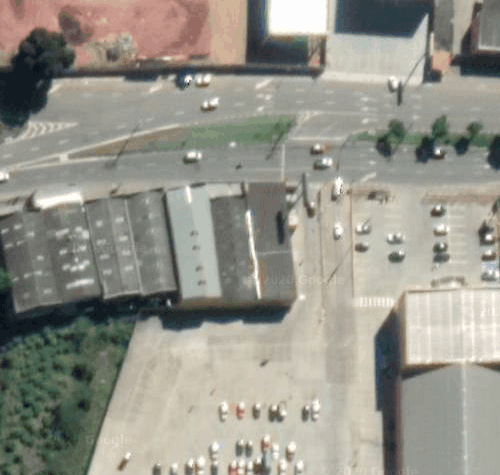} \\
    		\includegraphics[width=\exCV\textwidth, height=\exCV\textwidth]{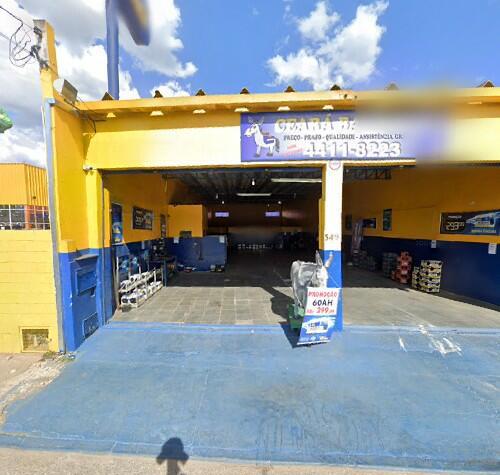}
        \end{tabular}
	}
	\caption{Examples of the CV-BrCT Dataset.}
	\label{fig:cv_brct}
\end{figure}


\subsection{Experimental Protocol} \label{subsec:protocol}



For both datasets, we employed a 5-fold cross-validation protocol, in which $80\%$ of the images are used for training, $10\%$ for validation, and the remaining $10\%$ for testing.
Following this protocol, we simulated and evaluated two different missing data scenarios using the test set: one for aerial and one for ground.


Considering this, the retrieval network (described in Section~\ref{subsec:impl}) was trained using the following hyper-parameters: $200$ epochs,
batch size equal to the number of classes of the dataset (i.e., $11$ for AiRound and $7$ for CV-BrCT), $\gamma$ (Equation~\ref{eq:soft_margin_triple_loss}) of $10$, Adam~\cite{kingma2014adam} as optimizer, learning rate of $0.00001$, and exponential decays of $0.9$ and $0.999$. 
Results related to this model are reported in terms of the average mean Average Precision at K (mAP@K)~\cite{valcarce2020assessing}, taken from all 5-fold experiments with its corresponding standard deviation.
%
Additionally, it is important to highlight that, during the inference, this network used the validation set as the retrieval database, i.e., images of the validation set are retrieved and then paired (based on their similarity) with the test samples for the final classification.
%

For the classification part, we evaluated three well-known architectures: VGG~\cite{vgg}, DenseNet~\cite{densenet}, and SKNet~\cite{sknet}.
All networks were fine-tuned (from the ImageNet~\cite{imagenet_cvpr09} dataset) for each of the domains (aerial and ground) using the following hyper-parameters: $200$ epochs, early stop with $20$ epochs, batch size of $32$, stochastic gradient descent as optimizer, a learning rate of $0.001$, and momentum of $0.9$.
In this case, all obtained results are reported in terms of the average F1-Score and standard deviation among all 5 folds.

\subsection{Baselines} \label{subsec:baseline}

Four techniques were considered as baselines for both
datasets.
The first baseline, referenced as ``No Fusion'', consists of using a single-view classification CNN evaluated in the data available in the test phase (without any completion).
The idea of this baseline is to establish a lower bound for the other experiments.
The second baseline, referenced hereafter as ``Fully-Paired'', performs final classification assuming that the test set has no missing data (i.e., that it is fully paired).
To do so, this baseline uses two classification CNNs to extract features from both aerial and ground domains which are then fused (using the product fusion presented in Equation~\ref{eq:prod}) to produce the final result.
As for the second baseline, the idea is to set an upper bound for further experiments.

The remaining baselines come from the literature.
One of those is the EmbraceNet~\cite{choi2019embracenet}, a multi-view framework that learns a multimodal distribution to select the most relevant features of each view.
This distribution can be adjusted depending on the availability or absence of certain domains, thus being able to handle missing data.
Based on such framework, we proposed a two-stream network in which the final layer is actually an EmbraceNet layer~\cite{choi2019embracenet}, capable of efficiently dealing with missing data.
Such network was trained using the same set of hyper-parameters used for the classification models (Section~\ref{subsec:protocol}).

The last baseline is the Canonical Correlation Analysis (CCA)~\cite{srivastava2019understanding}.
Such an approach learns matrices that project the features from different views into a common latent space, bringing closer the varying perspectives of the same object~\cite{srivastava2019understanding}.
Using such projection matrices (learned from the training data) we can project the available view of the test set into the common latent space and then retrieve the closest scene to finally fill the missing data gap.
Following the guidelines introduced in the original work~\cite{srivastava2019understanding}, we first project the features extracted from the last layer before the classification using a Principal Component Analysis (PCA), and then use them as input for the CCA~\cite{srivastava2019understanding}.


    \section{Results} \label{sec:results}

In this section, we present and discuss the obtained results.
Section~\ref{subsec:retrieval} presents the results related to the retrieval part whereas Section~\ref{subsec:classification} evaluates and compares different configurations for the classification part of the proposed framework.
Finally, Section~\ref{subsec:baselines} compares the proposed framework with state-of-the-art baselines.


\subsection{Retrieval Analysis} \label{subsec:retrieval}

\begin{figure*}[ht]%
    \centering
    \subfloat{
        \includegraphics[width=.95\columnwidth]{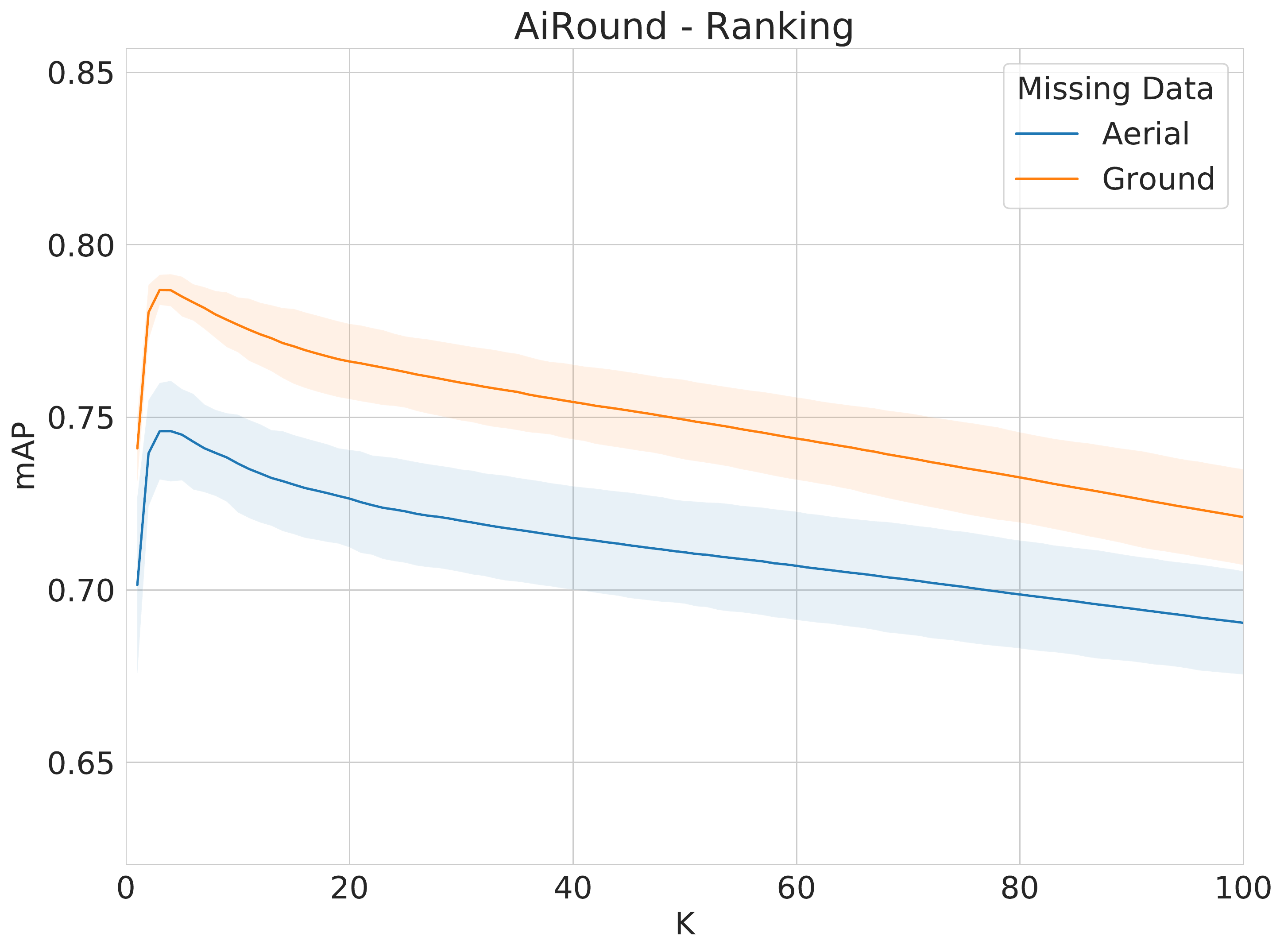}
    }%
    \qquad
    \centering
    \subfloat{
        \includegraphics[width=.95\columnwidth]{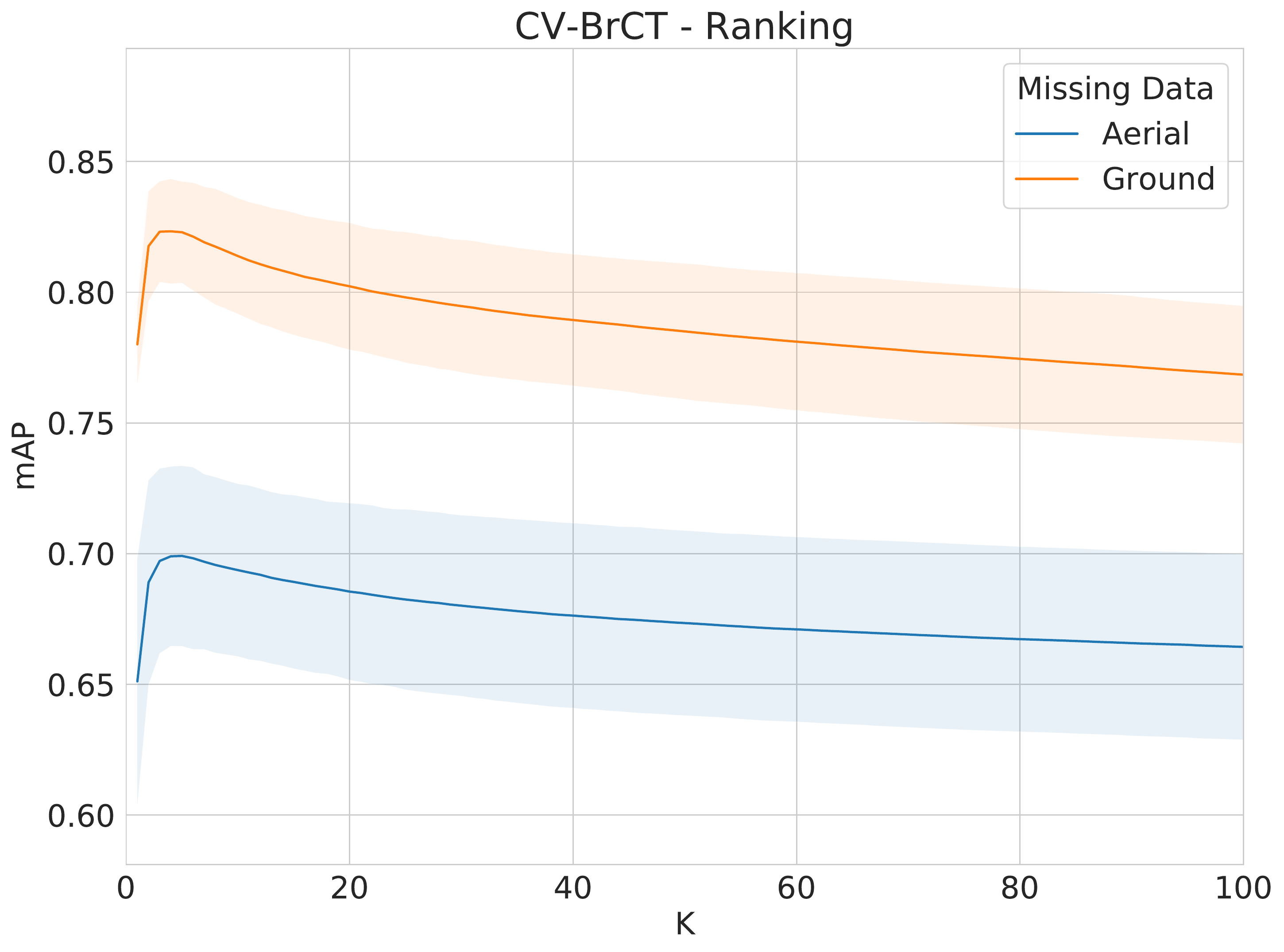}
    }%
    \caption{Results, in terms of mean Average Precision (mAP) and ranking size (K), of the retrieval network for the aerial and ground missing data scenarios in AiRound and CV-BrCT datasets.
    The shaded areas represent the standard deviation across the folds.}%
    \label{fig:results_ranking_size}%
\end{figure*}

In this Section, we analyze the retrieval network, a main component of the proposed framework.
Precisely, Figure~\ref{fig:results_ranking_size} reports the retrieval results, in terms of mAP@K~\cite{valcarce2020assessing}, for both AiRound and CV-BrCT datasets.


Analyzing the results, it is possible to observe that the retrieval network tends to produce better outcomes when using aerial data as input query (i.e., when the ground view is missing).
This may be justified by the fact that most aerial images tend to provide more context information than ground scenes to the retrieval model, that in turn exploits such useful data to recover similar images.
%
Another important aspect to discuss is the fact that, in general, the top-1 image does not produce the best results in terms of mAP.
This corroborates to our initial analysis about the potential of using top-k (instead of top-1) images to fill the missing data gap and, consequently, produce better classification results.
A better discussion about the use of top-k images to fill this missing data gap is presented in the next Section.
\subsection{Classification Analysis} \label{subsec:classification}

In this Section, we analyze the impact of the different classification networks (VGG~\cite{vgg}, DenseNet~\cite{densenet}, and SKNet~\cite{sknet}) and investigate the influence of the ranking size (i.e., top-k) in the final performance of the framework considering two distinct missing data scenarios (described in Section~\ref{subsec:protocol}), i.e., one for aerial and one for ground.

\begin{table}[ht]
\caption{Obtained results achieved by the proposed method for the AiRound dataset.}
\resizebox{\columnwidth}{!}{
    \begin{tabular}{@{}cclll@{}}
    \toprule
    \textbf{Ranking Size} & \textbf{Available Data} & \multicolumn{1}{c}{\textbf{VGG}~\cite{vgg}} & \multicolumn{1}{c}{\textbf{DenseNet}~\cite{densenet}} & \multicolumn{1}{c}{\textbf{SKNet}~\cite{sknet}} \\ \midrule
    \multicolumn{1}{l}{\textbf{Top 1}} & \multirow{8}{*}{\begin{tabular}[x]{@{}c@{}}Ground\\(Aerial Missing)\end{tabular}} & $0.74 \pm 0.01$ & $0.75 \pm 0.02$ & $0.75 \pm 0.02$ \\
    \multicolumn{1}{l}{\textbf{Top 2}} &  & $0.76 \pm 0.01$ & $0.76 \pm 0.02$ & $0.77 \pm 0.02$ \\
    \multicolumn{1}{l}{\textbf{Top 3}} &  & $0.76 \pm 0.01$ & $0.77 \pm 0.03$ & $0.77 \pm 0.02$ \\
    \multicolumn{1}{l}{\textbf{Top 4}} &  & $0.77 \pm 0.01$ & $0.77 \pm 0.02$ & $0.78 \pm 0.01$ \\
    \multicolumn{1}{l}{\textbf{Top 5}} &  & $0.77 \pm 0.01$ & $0.77 \pm 0.02$ & $0.78 \pm 0.01$ \\
    \multicolumn{1}{l}{\textbf{Top 10}} &  & $0.77 \pm 0.02$ & $0.77 \pm 0.02$ & $0.78 \pm 0.01$ \\
    \multicolumn{1}{l}{\textbf{Top 50}} &  & $0.77 \pm 0.02$ & $0.77 \pm 0.02$ & $0.78 \pm 0.00$ \\
    \multicolumn{1}{l}{\textbf{Top 100}} &  & $0.77 \pm 0.02$ & $0.77 \pm 0.02$ & $\mathbf{0.79 \pm 0.00}$ \\ 
    \midrule
    \multicolumn{1}{l}{\textbf{Top 1}} & \multirow{8}{*}{\begin{tabular}[x]{@{}c@{}}Aerial\\(Ground Missing)\end{tabular}} & $0.82 \pm 0.01$ & $0.83 \pm 0.01$ & $0.83 \pm 0.01$ \\
    \multicolumn{1}{l}{\textbf{Top 2}} & \multicolumn{1}{l}{} & $0.83 \pm 0.01$ & $0.84 \pm 0.02$ & $0.83 \pm 0.01$ \\
    \multicolumn{1}{l}{\textbf{Top 3}} & \multicolumn{1}{l}{} & $0.83 \pm 0.01$ & $0.84 \pm 0.02$ & $0.84 \pm 0.01$ \\
    \multicolumn{1}{l}{\textbf{Top 4}} & \multicolumn{1}{l}{} & $0.83 \pm 0.01$ & $0.84 \pm 0.02$ & $0.84 \pm 0.01$ \\
    \multicolumn{1}{l}{\textbf{Top 5}} & \multicolumn{1}{l}{} & $0.83 \pm 0.01$ & $0.84 \pm 0.02$ & $0.84 \pm 0.01$ \\
    \multicolumn{1}{l}{\textbf{Top 10}} & \multicolumn{1}{l}{} & $0.83 \pm 0.01$ & $0.85 \pm 0.02$ & $0.84 \pm 0.01$ \\
    \multicolumn{1}{l}{\textbf{Top 50}} & \multicolumn{1}{l}{} & $0.83 \pm 0.00$ & $0.84 \pm 0.02$ & $0.84 \pm 0.01$ \\
    \multicolumn{1}{l}{\textbf{Top 100}} & \multicolumn{1}{l}{} & $0.83 \pm 0.01$ & $\mathbf{0.85 \pm 0.01}$ & $0.84 \pm 0.01$ \\ \bottomrule
    \end{tabular}
}

\label{tab:results_airound}
\end{table}

Obtained results for the AiRound and CV-BrCT datasets are presented in Tables~\ref{tab:results_airound} and~\ref{tab:results_cv-brct}, respectively.
For both datasets and assessed scenarios, the strategy of combining information from multiple images to fill the missing data gap produced better outcomes than using just the top retrieved image (an approach commonly exploited in the literature~\cite{zhang2018multi,srivastava2019understanding}). 
This directly corroborates with our initial analysis about the potential of combining information from the top-k images to fill the missing data gap.

Aside from this, for the AiRound dataset, the best result for the aerial missing scenario was produced by the SKNet model~\cite{sknet} whereas the best outcome for the ground missing scenario was yielded by the DenseNet network~\cite{densenet}.
In both cases, the ranking size of $100$ (i.e., the top-100 images) yielded the best outcomes.
For the CV-BrCT dataset, all assessed models achieved very similar results when using a ranking size greater than or equal to 3 for the aerial missing scenario, and greater than or equal to 2 for the ground missing scenario.
Given this, the simplest model, i.e., VGG network~\cite{vgg}, with ranking size 100 was selected and used for further experiments using this dataset.

\begin{table}[ht]
\caption{Obtained results achieved by the proposed method for the CV-BrCT dataset.}
\resizebox{\columnwidth}{!}{
    \begin{tabular}{@{}cclll@{}}
    \toprule
    \textbf{Ranking Size} & \textbf{Available Data} & \multicolumn{1}{c}{\textbf{VGG}~\cite{vgg}} & \multicolumn{1}{c}{\textbf{DenseNet}~\cite{densenet}} & \multicolumn{1}{c}{\textbf{SKNet}~\cite{sknet}} \\ \midrule
    \multicolumn{1}{l}{\textbf{Top 1}} & \multirow{8}{*}{\begin{tabular}[x]{@{}c@{}}Ground\\(Aerial Missing)\end{tabular}} & $0.70 \pm 0.02$ & $0.70 \pm 0.03$ & $0.70 \pm 0.02$ \\
    \multicolumn{1}{l}{\textbf{Top 2}} &  & $0.72 \pm 0.02$ & $0.72 \pm 0.02$ & $0.72 \pm 0.02$ \\
    \multicolumn{1}{l}{\textbf{Top 3}} &  & $0.72 \pm 0.02$ & $\mathbf{0.73 \pm 0.02}$ & $0.72 \pm 0.02$ \\
    \multicolumn{1}{l}{\textbf{Top 4}} &  & $\mathbf{0.73 \pm 0.02}$ & $\mathbf{0.73 \pm 0.02}$ & $\mathbf{0.73 \pm 0.03}$ \\
    \multicolumn{1}{l}{\textbf{Top 5}} &  & $\mathbf{0.73 \pm 0.02}$ & $\mathbf{0.73 \pm 0.02}$ & $\mathbf{0.73 \pm 0.02}$ \\
    \multicolumn{1}{l}{\textbf{Top 10}} &  & $\mathbf{0.73 \pm 0.02}$ & $\mathbf{0.73 \pm 0.01}$ & $\mathbf{0.73 \pm 0.02}$ \\
    \multicolumn{1}{l}{\textbf{Top 50}} &  & $\mathbf{0.73 \pm 0.02}$ & $\mathbf{0.73 \pm 0.02}$ & $\mathbf{0.73 \pm 0.03}$ \\
    \multicolumn{1}{l}{\textbf{Top 100}} &  & $\mathbf{0.73 \pm 0.02}$ & $\mathbf{0.73 \pm 0.02}$ & $\mathbf{0.73 \pm 0.02}$ \\ 
    \midrule
    \multicolumn{1}{l}{\textbf{Top 1}} & \multirow{8}{*}{\begin{tabular}[x]{@{}c@{}}Aerial\\(Ground Missing)\end{tabular}} & $0.87 \pm 0.02$ & $0.87 \pm 0.02$ & $0.87 \pm 0.02$ \\
    \multicolumn{1}{l}{\textbf{Top 2}} & \multicolumn{1}{l}{} & $0.87 \pm 0.01$ & $\mathbf{0.88 \pm 0.02}$ & $0.87 \pm 0.02$ \\
    \multicolumn{1}{l}{\textbf{Top 3}} & \multicolumn{1}{l}{} & $\mathbf{0.88 \pm 0.01}$ & $\mathbf{0.88 \pm 0.01}$ & $0.87 \pm 0.02$ \\
    \multicolumn{1}{l}{\textbf{Top 4}} & \multicolumn{1}{l}{} & $\mathbf{0.88 \pm 0.01}$ & $\mathbf{0.88 \pm 0.01}$ & $0.87 \pm 0.02$ \\
    \multicolumn{1}{l}{\textbf{Top 5}} & \multicolumn{1}{l}{} & $\mathbf{0.88 \pm 0.01}$ & $\mathbf{0.88 \pm 0.01}$ & $0.87 \pm 0.02$ \\
    \multicolumn{1}{l}{\textbf{Top 10}} & \multicolumn{1}{l}{} & $\mathbf{0.88 \pm 0.01}$ & $\mathbf{0.88 \pm 0.02}$ & $\mathbf{0.88 \pm 0.01}$ \\
    \multicolumn{1}{l}{\textbf{Top 50}} & \multicolumn{1}{l}{} & $\mathbf{0.88 \pm 0.01}$ & $\mathbf{0.88 \pm 0.02}$ & $\mathbf{0.88 \pm 0.01}$ \\
    \multicolumn{1}{l}{\textbf{Top 100}} & \multicolumn{1}{l}{} & $\mathbf{0.88 \pm 0.01}$ & $\mathbf{0.88 \pm 0.02}$ & $\mathbf{0.88 \pm 0.01}$ \\ \bottomrule
    \end{tabular}
}
\label{tab:results_cv-brct}
\end{table}


\subsection{State-of-the-art Comparison} \label{subsec:baselines}

This Section compares and discusses the results obtained by the proposed framework and the state-of-the-art baselines.
Observe that:
(i) based on the analyses carried out in the previous Sections, only the best results for each dataset and scenario are presented.
(ii) the baselines were conceived using the same network selected for the classification part of the proposed framework.
Precisely, for the AiRound dataset, baselines are based on the SKNet model~\cite{sknet} (for the aerial missing scenario) and on the DenseNet~\cite{densenet} (for the ground missing scenario).
For the CV-BrCT, all baselines were conceived based on the VGG network~\cite{vgg}.
(iii) all results presented here were verified by a fold-by-fold paired t-test with confidence level of 95\%.

\begin{table}[ht]
\caption{Results of the proposed method and baselines for AiRound dataset.}
\resizebox{\columnwidth}{!}{
    \begin{tabular}{@{}lcl@{}}
    \toprule
    \multicolumn{1}{c}{\textbf{Method}} & \textbf{Available Data} & \multicolumn{1}{c}{\textbf{F1 Score}} \\ \midrule
    No Fusion (Lower bound) & \multirow{4}{*}{\begin{tabular}[x]{@{}c@{}}Ground\\(Aerial Missing)\end{tabular}} & $0.76 \pm 0.00$ \\
    EmbraceNet~\cite{choi2019embracenet} &  & $0.69 \pm 0.03$ \\
    CCA~\cite{srivastava2019understanding} &  & $0.75 \pm 0.02$ \\
    \textbf{Ours} &  & $\mathbf{0.79 \pm 0.00}$ \\     \midrule
    No Fusion (Lower bound) & \multirow{4}{*}{\begin{tabular}[x]{@{}c@{}}Aerial\\(Ground Missing)\end{tabular}} & $0.83 \pm 0.00$ \\
    EmbraceNet~\cite{choi2019embracenet} &  & $0.83 \pm 0.00$ \\
    CCA~\cite{srivastava2019understanding} &  & $0.83 \pm 0.01$ \\
    \textbf{Ours} &  & $\mathbf{0.85 \pm 0.01}$ \\      \midrule
    Fully-Paired (Upper bound) & Aerial + Ground & $0.91 \pm 0.01$ \\ \bottomrule
    \end{tabular}
}
\label{tab:results_airound_baseline}
\end{table}

Tables~\ref{tab:results_airound_baseline} and~\ref{tab:results_cv-brct_baseline} present the results for the AiRound and CV-BrCT dataset.
Overall, the proposed approach outperformed all baselines except, as expected, the upper bound one.
In fact, for the CV-BrCT dataset, the difference between the results achieved by the proposed framework and the fully-paired (upper bound) baseline for the ground missing scenario is almost irrelevant.
However, for all other scenarios and datasets, this difference is considerable, which shows that there is still room for improvements.

\begin{table}[ht]
\caption{Results of the proposed method and baselines for CV-BrCT dataset.}
\resizebox{\columnwidth}{!}{
    \begin{tabular}{@{}lcl@{}}
    \toprule
    \multicolumn{1}{c}{\textbf{Method}} & \textbf{Available Data} & \multicolumn{1}{c}{\textbf{F1 Score}} \\ \midrule
    No Fusion (Lower bound) & \multirow{4}{*}{\begin{tabular}[x]{@{}c@{}}Ground\\(Aerial Missing)\end{tabular}} & $0.72 \pm 0.01$ \\
    EmbraceNet~\cite{choi2019embracenet} &  & $0.53 \pm 0.04$ \\
    CCA~\cite{srivastava2019understanding} &  & $0.72 \pm 0.02$ \\
    \textbf{Ours} &  & $\mathbf{0.73 \pm 0.02}$ \\     \midrule
    No Fusion (Lower bound) & \multirow{4}{*}{\begin{tabular}[x]{@{}c@{}}Aerial\\(Ground Missing)\end{tabular}} & $0.86 \pm 0.01$ \\
    EmbraceNet~\cite{choi2019embracenet} &  & $0.77 \pm 0.03$ \\
    CCA~\cite{srivastava2019understanding} &  & $0.86 \pm 0.01$ \\
    \textbf{Ours} &  & $\mathbf{0.88 \pm 0.01}$ \\      \midrule
    Fully-Paired (Upper bound) & Aerial + Ground & $0.89 \pm 0.02$ \\ \bottomrule
    \end{tabular}
}

 \label{tab:results_cv-brct_baseline}
\end{table}


Aside from this, it is interesting to observe that, for both datasets, the obtained results for the aerial missing scenario are worse than the outcomes for the ground missing scenario.
As previously explained, this may be justified by the fact that most aerial images tend to provide more context information (than ground scenes), thus assisting in the classification process and, consequently, yielding better results.

    \section{Conclusions} \label{sec:conclusion}

In this paper, we propose a novel framework to handle multi-view image classification with missing data.
The proposed approach is composed of two main parts: (i) a retrieval one, responsible for recovering similar samples that can be used to fill the missing data gap of the input query image; and
(ii) a classification one, that fuses information extracted from both the input image and the top-k retrieved scenes in order to perform the final classification.

Experiments were conducted using two multi-view aerial-ground datasets (AiRound and CV-BrCT) and considering two distinct scenarios: one in which aerial images are considered absent and another where ground scenes are considered missing.
%
Results have showed that the proposed technique is efficient and robust.
Precisely, it achieved state-of-the-art results in both datasets outperforming all baselines, except for the fully paired (upper-bound) one.
Additionally, experimental outcomes showed that the strategy of combining information from multiple (i.e., top-k) images to fill the missing data gap (instead of using just the top retrieved image, as commonly employed in the literature~\cite{zhang2018multi,srivastava2019understanding}) is remarkably effective.

As future work, we intend to evaluate the proposed approach using other datasets with more (than two) views per instance.
We also would like to assess other retrieval methods as well as other classification networks.

    
    \section*{Acknowledgments}

    The authors acknowledge the Serrapilheira Institute (grant Serra – R-2011-37776).
    This work was supported in part by the Minas Gerais Research Funding Foundation (FAPEMIG) under Grant APQ-00449-17, 
    by the National Council for Scientific and Technological Development (CNPq) under Grant 306955/2021-0, and
    by the \emph{Coordenação de Aperfeiçoamento de Pessoal de Nível Superior – Brasil (CAPES)} – Finance Code 001.
    
    \bibliographystyle{elsarticle-num-names}
    \bibliography{refs}

\end{document}